\documentclass[11pt]{article}

\usepackage[final]{acl}
\usepackage{times}
\usepackage{latexsym}
\usepackage[T1]{fontenc}
\usepackage[utf8]{inputenc}
\usepackage{microtype}
\usepackage{inconsolata}
\usepackage{graphicx}
\usepackage{epigraph}
\usepackage{booktabs}
\usepackage{algorithm}
\usepackage{algpseudocode}
\usepackage{amsmath}
\usepackage{caption}
\usepackage{amsfonts}
\usepackage{tabularx}

\title{DARWIN: Dynamic Agentically Rewriting Self-Improving Network}

\author{Henry Jiang \\
\textit{College of Computing, Georgia Institute of Technology}\\
  \texttt{hjiang378@gatech.edu}}

\begin{document}
\maketitle

\section{Abstract}
\textbf{DARWIN is an evolutionary GPT model, utilizing a genetic-algorithm like optimization structure with several independent GPT agents being trained individually using unique training code. Each iteration, the GPT models are prompted to modify the training code of one another in an attempt to improve their performance in a mutation-like manner, and the best GPT agents are then benchmarked and selected for the next iteration by genetic algorithm. 
For demonstration purposes and due to budget and time constraints, OpenAI API is used to prompt training code improvements and the nanoGPT framework is used as the training code. DARWIN also utilizes persistent JSON-based memory files to track previous reasoning and changes to code to correlate with improvement to model performance. and a bidirectional interface for HITL intervention allowing the model to request upgrades such as additional datasets, training scripts, and restructuring of file hierarchies. In experiments, DARWIN achieved a 1.26 percent improvement in model FLOPS utilization (MFU) and a 2.07 percent improvement to perplexity in 5 iterations of training over baseline configurations, demonstrating promising capabilities as a foundation for scaling evolutionary GPT training.}

\section{Introduction}
Modern LLMs currently operate in the realm of artificial narrow intelligence, only being capable of performing specific tasks that they are tuned to. The pursuit of AGI continues to elude researchers at the frontier of modern AI understanding, and the concept is commonly intertwined with self-improvement of models. In fact, systems that recursively enhance their capabilities beyond the abilities and understanding of human programming have been theorized and proposed since GSchmidhuber's first proposal of Gödel machines based on Gödel's work on self-referencing mathematical theorems \citep{Schmidhuber2006Godel}. Modern LLMs remain bottlenecked in performance by their inability to tune to specific tasks and in their static state post-training, unable to autonomously refine and improve upon their architectures and training procedures. As the mathematics behind modern neural network and transformer architecture is difficult to interpret why it operates successfully, AI agents may have a more successful time interpreting their own performance bottlenecks and successful strategies and implementing them into future iterations of themselves, especially with the strong coding capabilities of LLM models.

This paper presents DARWIN (\textbf{D}ynamic \textbf{A}gentically \textbf{R}e\textbf{W}riting Self-\textbf{I}mproving \textbf{N}etwork), a framework that synthesizes evolutionary GPT concepts into a genetic algorithm model of selecting future models, utilizing a unique mutation substitute of prompting parent models to modify one another's training code. Careful containerization, memory file systems, and HITL interfaces for model requests supplement the system to provide the framework with further options to supplement prompts and allow the system to evolve beyond bottlenecks like file tree hierarchy and dataset sizes.

The contributions are as follows:
\begin{itemize}
    \item DARWIN, an open-source framework for training evolutionary GPT optimization of training code.
    \item Demonstrate proof-of-concept results with nanoGPT \cite{Karpathy2022} as the framework of choice.
    \item Present architectural innovations on persistent memory storage across iterations and bidirectional HITL interfaces.
    \item Discuss implications for optimizing high computational and time requirement bottlenecks as pathway for scaling AGI.
\end{itemize}

\section{Related Work}
Development of self-improving AI draws from several fields, including evolutionary machine learning, large language model research, and automata theory. This section examines the foundations for evolutionary LLM reasoning and recursive improvement and recent advances and works in frameworks.

\subsection{Foundational Theories}
Formal study of self-improving machine learning algorithms can trace to Schmidhuber's Gödel machine proposal \cite{Schmidhuber2006Godel}, a theoretical framework combining Darwin's evolution theoretical frameworks with self-referencing real theorems proposed by Kurt Gödel in the 1930s. Gödel machines operate on a dual process of an incumbent policy executing tasks while a searcher component searches for beneficial modifications. This architecture in practicality is limited by computational intractibility of generating proofs that guarantee improvement for complex systems like neural networks.

Evolutionary computation and LLMs intersect in that LLMs can serve as effective mutation and crossover operations \cite{lehman2022evolutionlargemodels}. this establishes that the vast dataset used to train LLMs implicitly encodes rich priors that enable semantically meaningful variations akin to random perturbations. Furthermore, Self-Taught Optimizer (STOP) was a framework proposed that demonstrated recursively self-improving code generation \cite{zelikman2024selftaughtoptimizerstoprecursively}. STOP iteratively refines its own optimization algorithms and generates potentially improved versions. It then validates on benchmark problems, seeking to maximize optimization via Gödel machine-like principles.

\subsection{Agentic Evolutionary Workflows}
Recent work has explored practical implementations of recursive self-improvement frameworks enabling LLM agents to modify logic and reasoning procedures \cite{yin2025godelagentselfreferentialagent}. The Darwin-Gödel Machine \cite{zhang2025darwingodelmachineopenended} implements a successful proof of concept using LLM API calls for iterative improvement through beam search as global optimizer. DARWIN shares many alignments with the approach and takes inspriation on the task of training code optimization with a focus on safe containerization and automated iterative improvement. 

Other notable works include EvoAgentX \citep{wang2025evoagentx} which provides an automated platform for testing multi agent systems and optimizing agentic workflows, allowing the testing of a multitude of model setups. More pertinent to the system of persistent memory of previous modifications and agent modifications, HealthFlow \citep{zhu2025healthflowselfevolvingaiagent} establishes a robust JSON file memory system for meta planning capabilities for autonomous healthcare tasks, checking for successful implementation and automatically resolving issues in code.

DARWIN synthesizes insights from these diverse works and concepts across theoretical fields to enable learning across evolutionary generations. Fault tolerant debugging, containerization, and supplementary tools for the agent allow the genetic algorithm-like framework to operate safely and effectively, bridging theoretical self-improvement frameworks and deployable LLMs. 

\section{Design}

\subsection{Architecture Overview}
DARWIN implements evolutionary optimization with four modules: a central controller for genetic algorithm generations and training multiple models in parallel, a mutation script for prompting modification of the code of pairs of LLM agents, a fitness evaluation step with benchmarking, and a utilities script for memory and the bidirectional HITL communication interface.

\begin{figure}[h]
    \centering
    \includegraphics[width=0.5\textwidth]{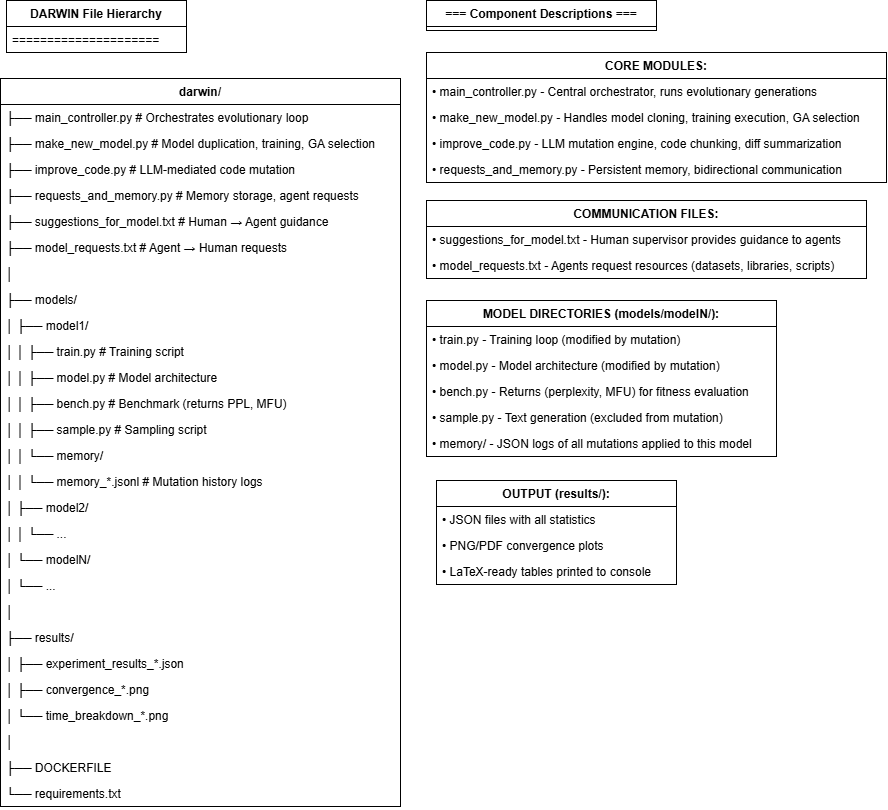}
    \caption{DARWIN File Hierarchy and Relations}
    \label{fig:my_label}
\end{figure}

\subsection{Containerization}
Due to the nature of DGMs in modifying source code, it is pertinent that these models operate in individual containers and proper isolation is carefully considered and monitored during design and execution \citep{zhang2025darwingodelmachineopenended}. In DARWIN, the Dockerfile system contains the central control script and all individual agents contained in seperate folders. Furthermore, DARWIN should be run on a containerized virtual machine as well. The prompt provided to the LLM agent explicitly states not to provide root access to the directory containing the central control script. Currently, manual checks for potentially damaging behavior are performed.

If moving beyond the proof of concept stage, it is inherent to upgrade the sophistication beyond manual checks, incorporating automated detection of unwanted behavior in the central control script to prune agents before the training loop begins. The Darwin-Gödel Machine utilizes a system of creating individual Dockerfiles for each agent to train, then subsequently retrieving results and destroying the temporary container \cite{zhang2025darwingodelmachineopenended}. This takes a long time, especially if scaled to large LLMs with massive datasets, but it may be a necessary ethical precaution for potentially unwanted and dangerous behavior if left completely unmonitored.

\subsection{Evolutionary Loop}

\begin{figure}[h]
    \centering
    \includegraphics[width=0.5\textwidth]{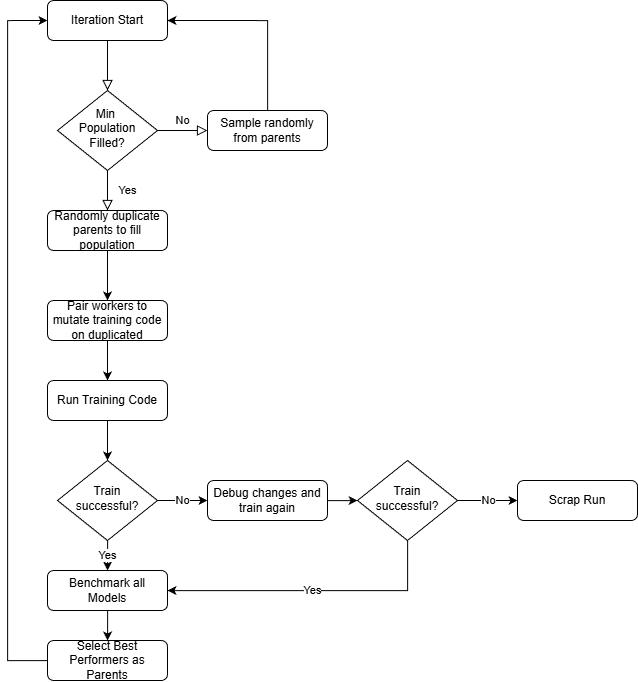}
    \caption{Flow Chart of Full Evolutionary Loop}
    \label{fig:my_label}
\end{figure}

The central controller orchestrates the evolutionary process in iterations, performing several steps. To begin, if the number of agents is below the population count which can occur if past models failed to run properly, the central controller samples randomly from the existing agents to fill the population to have enough parents for the crossover and mutation step. 

Next, offspring directories are randomly paired to worker processes and each worker is prompted to mutate the Python files in the offspring directory with the intention of improving the outcome of the final trained model. Due to constraints, this is currently delegated to an OpenAI API call. This excludes evaluation scripts for sampling and benchmarking; this check could be made more robust in the future. This process is repeated until there are enough new agents in the population to begin training.

Next, each offspring is then trained in an isolated working directory in parallel and benchmarked with a standardized benchmark. Currently due to the simple nature of nanoGPT using a Shakespeare prose dataset, the benchmarks of choice are MFU and perplexity. In the event that a model fails to train, the worker process is provided one opportunity to correct the potential errors in the modified script. This model is removed from consideration as a survivor for the next generation if it fails to train again. 

After all models are trained, a subset of the best performers are selected for the following generation and the loop repeats.
\subsection{Prompting Strategy and Mutation}

The prompting strategy for mutating source training code requires more sophistication to modify a complete complex system, maintaining a running memory of file hierarchy and previous structure when the max prompt token size is not enough to contain the entire script file. Source files are separated into individual chunks for module-level imports, class definitions, and function bodies using regex extraction. 

Three types of data supplement the prompt for mutation. Global variables, file structure data, and summaries of previously modified chunks is maintained in running memory to allow modification of individual chunks while preserving understanding of overall structure and minimizing the chance of syntax based errors. Each chunk is prompted for modification with a set probability to simulate mutation; module level imports are unchanged to preserve dependency consistency, although these can be modified upon request by the model for human intervention using the bidirectional communication interface discussed later. Overall, prompts direct the agent to modify with performance, code cleanliness, and training speed all in mind.
 
\subsection{2-Way Interface and Memory}

DARWIN supplements the training process with additional features to enable agent requests and human guidance to address potential bottlenecks in training dataset size, reorganization of file hierarchy to maintain clean code structure, additional module level imports and library dependencies, and additional scripts for training. This in practice would allow a human operator to check between iterations or several iterations for individual modification requests and decide for each one whether or not to fulfill the request. This interface also allows the human operator to inform the agent of the approved modifications syntactically, which is also provided as supplementary information in the prompt for training code modification/mutation.

\begin{figure}[h]
    \centering
    \includegraphics[width=0.5\textwidth]{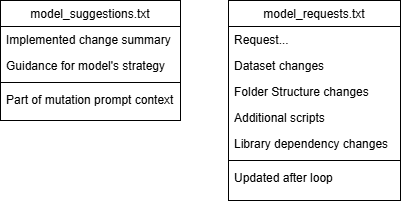}
    \caption{Bidirectional Communication Interface Diagram}
    \label{fig:my_label2}
\end{figure}

DARWIN also maintains a history of past evolutionary history through JSON files similar to Healthflow \cite{yin2025godelagentselfreferentialagent}. This records past modifications to the source code and the correlates to corresponding results on benchmark performance to guide the agent in future decisions. Each mutation event generates a record containing timestamp, source file path, LLM-generated summary, token counts and backend metadata. 

\section{Experimental Evaluation}
\begin{table*}[t]
\centering
\caption{Performance metrics across evolutionary generations.}
\label{tab:evolution}
\begin{tabular}{ccccccc}
\hline
Gen & Best PPL & Mean PPL & Std PPL & Best MFU (\%) & Mean MFU (\%) & Time (s) \\
\hline
0 & 38.05 & 38.32 & 0.17 & 0.40 & 0.40 & 474.0 \\
1 & 38.29 & 38.48 & 0.12 & 0.40 & 0.40 & 27.8 \\
2 & 38.68 & 38.93 & 0.15 & 0.40 & 0.39 & 300.7 \\
3 & 37.70 & 38.38 & 0.53 & 0.41 & 0.40 & 112.6 \\
4 & 38.57 & 38.80 & 0.18 & 0.40 & 0.39 & 199.8 \\
\hline
\end{tabular}
\end{table*}
DARWIN's capabilities are evaluated by using nanoGPT as the framework and substituting model calls for OpenAI API calls instead due to constraints on time and computation resources. The LLM backend utilized was GPT-4o-mini. Baseline configuration consisted of population of 10 separate models and retaining only 4 selected via GA. Model training hyperparameters consisted of 100 iterations, batch size of 64, and block size of 256 with a simple 6 layer, 6 head, 384 embedding model. Dropout was 0.2 and learning rate was 1e-3,  DARWIN was run for 5 generations with a mutation probability of 0.3, rather high due to the low number of total chunks for nanoGPT.Table 1 summarizes performance across evolutionary generations.

Baseline perplexity was 38.498 and baseline MFU was 0.397 After 5 generations DARWIN achieved a best perplexity of 37.70 and a best MFU of 0.392. This marks a 2.07 percent reduction in perplexity and a 1.26 percent reduction in MFU. Average time taken per generation was 223 seconds. Total error statistics were also tracked; across 50 total training instances, 18 errors were encountered of which 3 were resolved. This provides us an error rate of 37.5 percent and an error resolution rate of 16.67 percent.

Overall, results were not very promising and there is no significant improvement by the LLM modifications. However, it must be noted that this is comparing an optimized model to another optimized model on a simple task of generating Shakespeare prose. It is still hard to tell the effectiveness of such a strategy with such simple training code and scenario and so few iterations of improvement. 

\begin{table}[h]
\centering
\caption{Training Hyperparameters}
\label{tab:hyperparams}
\begin{tabular}{ll}
\toprule
\textbf{Hyperparameter} & \textbf{Value} \\
\midrule
Baseline Perplexity & 38.49836587905884 \\
Baseline MFU & 0.39698676646763204 \\
Final Best Perplexity & 37.69668039284751 \\
Final Best MFU & 0.3922781358644053 \\
Perplexity Reduction & -0.18 \\
MFU Improvement & -0.18 \\
Total Training Instances & 48 \\
Total Errors & 18 \\
Resolved Errors & 3 \\
Error Rate & 37.50 \\
Error Resolution Rate & 16.67 \\
Avg Time/Generation & 223.0 seconds \\
Total Time & 1114.8 seconds \\
\bottomrule
\end{tabular}
\end{table}

\subsection{Ablation Study}
Studies were also conducted on performance without the memory file system. Removing persistent memory resulted in a 3 percent worse performance compared to without the memory system, although it is hard to extrapolate these results to meaningful conclusions with insignificant results.

\subsection{Model Requests}
The following is an example of one of the descriptions in a memory file used as reference for future iterations of model improvement:\\
\begin{verbatim}
{**Improved Memory Management**:
- The `get_batch` function now uses a 
single `np.memmap` call for both training
and validation datasets, reducing redundancy
and potential memory overhead.
**Enhanced Loss Calculation**: 
- The `estimate_loss` function now 
initializes the `losses` tensor directly on 
the specified device, which can improve 
performance by avoiding unnecessary data 
transfers between CPU and GPU.}
\end{verbatim}

The model effectively summarizes all modifications to source code and the intended effect of the modification. These memory files would theoretically allow the model to make future decisions with past context and keep track of changes between versions. It also allows post-hoc analysis of what changes the model made that were effective in improving performance.

\section{Future Work}

DARWIN's current implementation serves as a proof-of-concept for foundational evolutionary GPT training for training code optimization. Many avenues are identified for future development to enhance the framework's capabilities and practical applicability. With a more complex model, better benchmarks for linguistic, mathematics and coding tasks could be implemented; this is crucial, as the benchmark must not be modified as to prevent the AI agent to tailoring the benchmark to exaggerate performance with confirmation bias. 

Scaling nanoGPT to GPT-2 and beyond level transformer architectures with billions of parameters is also a natural step for DARWIN, although this would require a great deal of resources to scale testing to this size. As a proposal to address potential concerns with high level of resource consumption per iteration, a distributed architecture system utilizing grid systems or GPU clusters for each individual agent with a central control system expanding on the present DARWIN framework could be implemented to parallelize the tasks and accelerate the time taken to train many GPT agents iteratively. Overall, this project opens up many options for furthering research and exploration of Darwin-Gödel machine architecture and sets the groundworks for future experimentation with the concept.

Source code can be viewed on Github at: https://github.com/henryyjiang/DARWIN

\bibliography{custom}

\end{document}